\documentclass[lettersize,journal]{IEEEtran}
\usepackage{amsmath,amsfonts}
\usepackage{algorithmic}
\usepackage{array}
\usepackage[caption=false,font=normalsize,labelfont=sf,textfont=sf]{subfig}
\usepackage{textcomp}
\usepackage{stfloats}
\usepackage[hyphens]{url}
\usepackage{bm}
\usepackage{amsmath, amssymb}
\usepackage{verbatim}
\usepackage{multirow}
\usepackage[dvipdfmx]{graphicx}
\def\BibTeX{{\rm B\kern-.05em{\sc i\kern-.025em b}\kern-.08em
    T\kern-.1667em\lower.7ex\hbox{E}\kern-.125emX}}
\usepackage{balance}
\usepackage{color}
\usepackage{jumoline}

\newcommand{\update}[1]{{#1}}
\newcommand{\delete}[1]{}
\newcommand{\moved}[1]{{#1}}

\begin{document}
\title{Easy Ensemble: Simple Deep Ensemble Learning for Sensor-Based Human Activity Recognition}
\author{Tatsuhito Hasegawa, \it{Member, IEEE},  Kazuma Kondo
\thanks{This work was supported in part by the Japan Society for the Promotion of Science (JSPS) KAKENHI Grant-in-Aid for Early-Career Scientists under Grant 19K20420.}}

\markboth{Journal of \LaTeX\ Class Files,~Vol.~18, No.~9, September~2020}%
{How to Use the IEEEtran \LaTeX \ Templates}

\maketitle

\begin{abstract}
  Sensor-based human activity recognition (HAR) is a paramount technology in the Internet of Things services. HAR using representation learning, which automatically learns a feature representation from raw data, is the mainstream method because it is difficult to interpret relevant information from raw sensor data to design meaningful features. Ensemble learning is a robust approach to improve generalization performance; however, deep ensemble learning requires various procedures, such as data partitioning and training multiple models, which are time-consuming and computationally expensive. In this study, we propose Easy Ensemble (EE) for HAR, which enables the easy implementation of deep ensemble learning in a single model. In addition, we propose various techniques (input variationer, stepwise ensemble, and channel shuffle) for the EE. Experiments on a benchmark dataset for HAR demonstrated the effectiveness of EE and various techniques and their characteristics compared with conventional ensemble learning methods.
\end{abstract}


\section{Introduction}
\IEEEPARstart{W}{ith} the widespread of Internet of Things (IoT) devices, sensor values can be easily collected and applied to human activity recognition (HAR\footnote{All acronyms are listed in Appendix.}). IoT sensors are appropriate for recording the behavior of individuals who move around compared to a fixed camera because the current IoT device enables wireless sensing. IoT devices enable the digitization of various natural information through sensors and wireless communication. IoT devices can be applied to many fields, such as monitoring a manufacturing process \cite{Muhammad2018}, and flood severity prediction \cite{Mohammed2020}. HAR is one of the vital applications of IoT technology. Nakamura et al. \cite{Nakamura2017} developed a general-purpose IoT sensing device: SenStick, and equipped it with some commodities, such as a pencil, chopsticks, and toothbrush. By using such divices, we can easily implement IoT sensing commodities; therefore, various sensor values are now continuously recorded by IoT devices. In contrast, it is difficult for humans to interpret the context of raw sensor values and human behavior differs among individuals; therefore, automatic feature extraction by deep representation learning is used to build HAR models. To develop an accurate HAR model, researchers have investigated various model architectures \cite{Xu2019_b, Long2019, kwang2019} using deep learning.

Because IoT devices can be easily equipped with many places, multiple sensors are generally used as inputs for the HAR models. For example, the well-known public benchmark HAR dataset PAMAP2 \cite{Reiss2012a} uses a heart rate sensor and three wireless inertial measurement units (IMUs). The IMUs are equipped at the wrist, chest, and ankle to measure temperature, three-axis acceleration (two scales), three-axis gyro, and three-axis geomagnetism. Although the shape of an input tensor in image recognition is commonly fixed to three RGB channels, many HAR studies use multidimensional sensor data as input. HAR sometimes applies ensemble learning \cite{Yao2017, Radu2018} to assign independent feature extractors for each sensor modality. Separate feature extractors for each sensor modality are generally superior to a single feature extractor that uses concatenated input data. As an advanced technique, a method that performs masking based on a multinomial distribution when merging each feature representation \cite{Choi2019}  and another that ensembles feature extractors by grouping them based on sensor features \cite{Chen2021} have also been proposed. Thus, the use of ensemble learning is more crucial in HAR than image recognition.

Ensemble learning is a simple strategy that can significantly enhance generalization performance, not only for HAR but also for various machine learning tasks. Ensemble learning is a general term for machine learning methods that use multiple models together, and it generally improves generalization performance compared with a single model. Ensemble learning is also a robust technique in deep learning. However, the computation and implementation costs are higher than those of a single model because multiple models must be trained individually, and then the models must be combined during inference to perform different behaviors than during training. Although a technique to reduce the computational cost by sharing model parameters \cite{Wasay2020} has been proposed, its implementation is complicated.

\begin{figure}[b]
    \begin{center}
    \includegraphics[width=7.0cm]{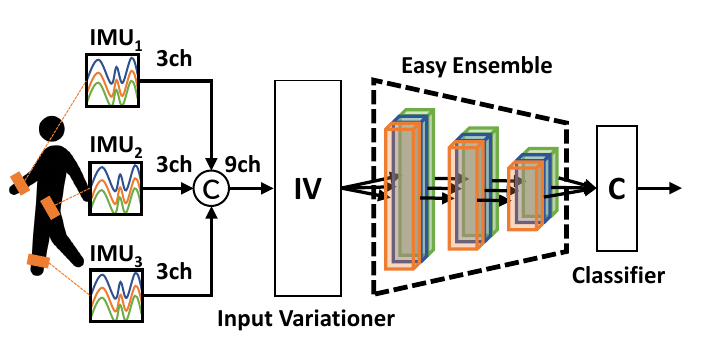}
    \caption{EE and input variationer make it possible to easily use a deep ensemble learning model for HAR.}
    \label{fig:abstract}
    \end{center}
\end{figure}

In this study, we propose a simple and easy-to-implement deep learning ensemble method --Easy Ensemble (EE)-- to improve the complexity of training and inference (Fig. \ref{fig:abstract}). EE is a method that treats the input channels as $N$ groups internally, each of which performs feature extraction independently. Particularly, we aim to realize an easy-to-implement deep ensemble learning method and not to reduce computational cost. We employ several benchmark datasets to demonstrate the effectiveness of EE and clarify factors contributing to the improvement of EE accuracy and characteristics of EE as each parameter or input is varied. Further, we propose different techniques: a new input diversification method for EE --input variationer--, parameter reduction by gradually changing the number of ensembles --stepwise ensemble--, and intersecting channels --channel shuffle--. We demonstrate its effectiveness in improving the generalization performance. \moved{The novelty of this study is that it focuses on realizing a method that is easy to implement and use rather than on reducing training time and memory cost.
}

\moved{The contributions of this study are as follows:}

\begin{itemize}
    \item \moved{We proposed a transformation algorithm, which enables various model architectures to be an ensemble model in a single model, and showed that EE is equivalent to the ensemble method with majority voting in the theoretical aspect.}    \item \moved{We proposed different techniques (input variationer, stepwise ensemble, and channel shuffle) for the EE. }
    \item \moved{Using a benchmark dataset for HAR, we clarified the characteristics of EE under various conditions. In addition, we found that EE's remarkable characteristics when the number of ensembles $N$ is large. One is that EE outperformed pure ensemble, and the other is that an ensemble model with fewer filters effectively work.}
\end{itemize}

\moved{The aim of this study is not to propose a new model architecture that improves estimation accuracy and inference speed but to propose a transformation algorithm of EE. We targeted a use case where an ensembling technique improves the estimation accuracy of a non-ensemble model. The idea of EE was proposed in ResNeXt \cite{Xie2017} and group ensemble (GE) \cite{chen2020group}, but this EE generalizes the ``Cardinality'' concept in ResNeXt, making it applicable to most model architectures. The paper of GE did not provide specific algorithms and proofs from a theoretical perspective. Therefore, this study's novelty is to clarify the generalization of ``Cardinality,'' details the transformation algorithm, and provides theoretical proofs. In the following sections, we described the related work and EE algorithm in Section II and III, respectively. Next, we evaluated EE through experimental results and clarified its characteristics in Section IV. Finally, we provided additional techniques and discussion in Section V-VI.}

\section{Related work}
\update{In the following sections, we describe studies related to the ensemble- and deep-learning-based HAR and the modern techniques of deep ensemble learning. We summarized them in Table \ref{table:related}. }
\begin{table*}[tb]
  \newcolumntype{A}{>{\centering\arraybackslash}p{2em}}
  \caption{Summary of the related work}
  \label{table:related}
  \hbox to\hsize{\hfil
  \begin{tabular}{lcccl} \hline \hline
Reference & Target & DL & Ensemble & Note \\ \hline
\cite{Subasi2018, Irvine2020, Nurul2021, Xu2019} & HAR &  & $\checkmark$ & Conventional machine learning with hand-crafted features. \\
\cite{Zhu2019, Semwal2021} & HAR & $\checkmark$ & $\checkmark$ & Practical example of deep ensemble learning. \\
\cite{Zeng2014, Ronao2015, Hammerla2016} & HAR & $\checkmark$ &  & Deep learning with hyperparameter tuning. \\
\cite{Ma2019, Zeng2018, Zhao2022} & HAR & $\checkmark$ &  & Deep learning architecture specified to HAR. \\
\cite{Wasay2020, lee2015m, Wen2020, Havasi2021} & Image & $\checkmark$ & $\checkmark$ & Model compression in deep ensemble learning. \\
 \cite{Wasay2021, Zhu2021} & Image & $\checkmark$ & $\checkmark$ & Analysis of trend and dynamics in deep ensemble learning. \\
\cite{chen2020group}  & Image & $\checkmark$ & $\checkmark$ & Deep ensemble learning in a single model. \\ \hline
  \end{tabular}\hfil}
\end{table*}

\subsection{Inference method in HAR}
There are many types of research on HAR using IoT sensors. The inference method of HAR can be broadly classified into conventional machine learning and deep learning. The most crucial difference between them is whether or not feature vectors are automatically designed. Each of them can use an ensemble approach to enhance the generalization performance.

In conventional HAR, many studies proposed various feature vectors designed manually and adopted ensemble learning algorithm as a classifier \cite{Subasi2018, Irvine2020, Nurul2021, Xu2019}. These methods only use multiple classifiers but use the same feature vectors for each classifier.

In contrast, deep-learning-based HARs use multiple feature extractors and classifiers. 
Zhu et al. \cite{Zhu2019} proposed a method that builds multiple classifiers using hand-crafted features, then uses convolutional neural networks (CNNs) to learn feature representations from raw data, and determines the final output using weighted voting. Semwal et al. \cite{Semwal2021} proposed a method to combine the outputs obtained using four models in parallel, each of which is a CNN connected with long short-term memory (LSTM), and then obtain the final output through a fully-connected layer. In most methods that do not use deep learning, the final output is determined by majority vote, weighted sum, or rule-based method from the outputs of multiple models. When deep learning is used, the acquired feature representations can be combined and connected to a new fully-connected layer, as described by Semwal et al. \cite{Semwal2021}
 
Because deep learning has various hyperparameters and model architectures, many studies investigated adequate conditions. 
Zeng et al. \cite{Zeng2014} examined the effect of hyperparameters in deep-learning-based HAR. Ronao et al. \cite{Ronao2015} proposed hyperparameter-tuned CNN in which the number of layers and feature maps were tuned using a greedy algorithm. Hammerla et al. \cite{Hammerla2016} also investigated various hyperparameters for CNN- and RNN-based HAR. Deep learning model architectures have also been studied in HAR. Ma et al. \cite{Ma2019} proposed a model combining CNN, RNN, and an attention technique. Zeng et al. \cite{Zeng2018} proposed temporal and sensor attention in LSTM to capture time series information in data. Our previous studies \cite{Zhao2022} have investigated which CNN architecture proposed in the image recognition field was better for sensor-based HAR. Our results indicated that Inception-v3 outperformed other architectures. However, the attention modules sometimes did not work efficiently due to the backbone architecture. Many studies have been conducted to achieve state-of-the-art HAR, and ensembling them can improve estimation accuracy.

\subsection{Deep ensemble learning method}
\subsubsection{Deep ensemble architecture}
Although ensemble learning improves the generalization performance of models, the training time and memory cost required for one model are high in deep learning. Several methods have been proposed to mitigate this issue in image recognition.

Lee et al. \cite{lee2015m} experimentally demonstrated an effective deep ensemble method. From the results, they proposed a method of ensembling while suppressing the number of parameters using TreeNets, a tree-structured model in which the lower layers share parameters, but the upper layers do not. It is preferable to merge values before the softmax function is applied instead of values that imitate the probability distribution after the softmax function is applied.

Wen et al. \cite{Wen2020} proposed a Batch Ensemble, which suppresses the number of parameters by representing multiple models to be ensembled as a Hadamard product of shared parameters ${\bm W}$ and a trainable rank-1 matrix ${\bm F_i}$. Similarly, Wasay et al. \cite{Wasay2020} proposed MotherNets, which shares the parameters of common substructures in a model. Models with multiple architectures were classified by clustering, and a common substructure (MotherNet) was manually constructed within each cluster. MotherNet is first pretrained and then restored to its original structure by adding a small noise to the trained parameters. Subsequently, each model is additionally trained, which enables the fast training of ensemble models.

Havasi et al. \cite{Havasi2021} stated that although some ensemble methods based on parameter sharing reduce memory cost, their inference speed is decreased due to multiple forward paths. They proposed multi-input multi-output, which improves ensemble performance by increasing the number of inputs and outputs while sharing the model itself.

Because image recognition uses massive high-resolution data, the training time and memory cost are significant issues. Moreover, in the HAR field, the input data are waveform data and training datasets are not as large as those of image recognition. Therefore, it is necessary to find a method to facilitate the implementation of ensemble learning to improve accuracy.

\subsubsection{Analysis on deep ensemble learning}
Although ensemble learning contributes to improving estimation accuracy because of its diversification, these exist some unexplained aspects. Wasay et al. \cite{Wasay2021} experimentally analyzed the performance of CNNs in image recognition, comparing an ensemble of multiple models with few parameters to a single model with the same number of parameters. By comparing several image recognition benchmarks, they found that ensemble learning needs to have a certain number of parameters to be effective and that the more complex the dataset is, the more effective the ensemble learning will be.

Allen-Zhu et al. \cite{Zhu2021} reported that increasing estimation accuracy using a deep ensemble learning and knowledge distillations \cite{Hinton2015} is a mystery\footnote{Microsoft Research Blog: Three mysteries in deep learning: Ensemble, knowledge distillation, and self-distillation [\url{https://www.microsoft.com/en-us/research/blog/three-mysteries-in-deep-learning-ensemble-knowledge-distillation-and-self-distillation/}]}. In an experiment using CIFAR-100, the average test accuracy of 10 individual models was 81.51\%, but the ensemble of weighted averaging of the output of each model improved the test accuracy to 84.69\%. However, training all models together by optimizing the sum of each model did not produce such an improvement in accuracy, resulting in a test accuracy of 81.83\%. Thus, in ensemble models with the same initial values, training the models aggregating outputs does not provide the ensemble benefit. In other words, individual training contributes to the acquisition of diversity rather than the effect of larger models.

\subsection{Ensemble learning with group convolution}
The novelty of our proposed method is that it uses grouping modules (group convolution, etc.) to build an ensemble characteristic in a single model. Group convolution has been used in the early years of deep learning. AlexNet has multiple CNN paths for parallel processing of multiple convolutions on multiple graphics processing units. In addition, ResNeXt \cite{Xie2017} introduces the concept of ``Cardinality,'' which splits convolution in the residual block of ResNet \cite{He2016} into multiple convolutions to improve accuracy. ``Cardinality'' can be implemented using multiple convolution layers, which is also equivalent to using group convolution.
\begin{figure*}[t]
    \begin{center}
    \includegraphics[width=16.0cm]{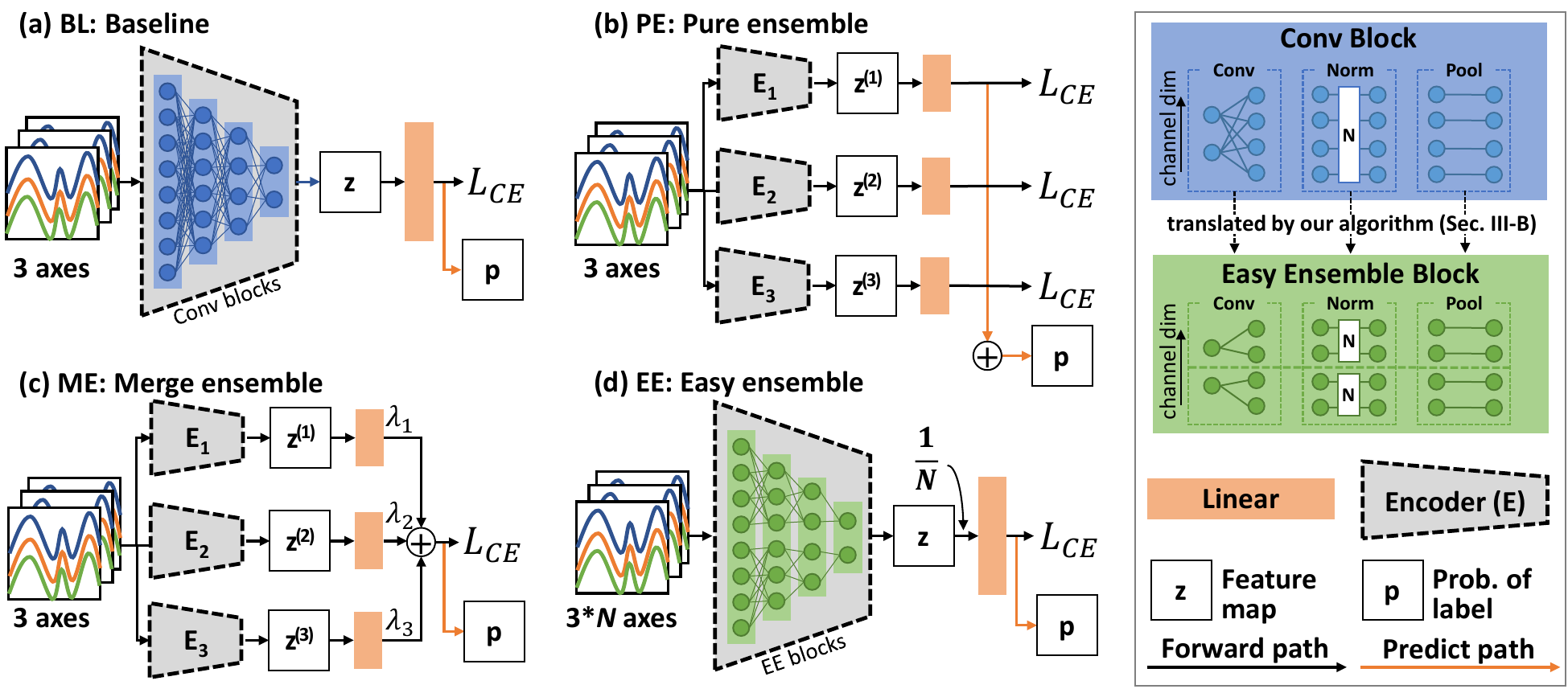}
    \caption{Model architecture of our proposed EE method and common ensemble of deep learning.}
    \label{fig:models}
    \end{center}
\end{figure*}

The method most similar to EE is the GE proposed by Chen et al. \cite{chen2020group}, which uses group convolution to make an ensemble learning in a single model for image recognition. Particularly, Chen et al. \cite{chen2020group} proposed sharing low layers and using random weights to combine multiple outputs. However, their algorithm was not described in detail. Moreover, there is no discussion on the equivalence before and after ensembling, normalization method, number of parameters, ablation study, and so on. There is also no discussion on input diversification because the lower layers are shared.

\section{EE}
\subsection{Ensemble models in deep learning}
Fig. \ref{fig:models} shows model outlines of EE and the general ensemble of deep learning methods. As shown in Fig. \ref{fig:models} (a), raw sensor data are input into the feature extractor --Encoder (E)-- in the general CNN model, and an extracted feature vector $\bm{z}$ is transformed into the probability distribution of activity label $\bm{p}$ through a fully-connected layer and the softmax function (hereinafter baseline (BL)). The parameters of the entire model were trained based on the categorical cross-entropy loss using $\bm{p}$ with a correct label.

Although various ensemble methods have been proposed for deep learning [Fig. \ref{fig:models} (b)], bagging-like ensemble is commonly used (hereinafter, pure ensemble (PE)). Assuming that the number of ensembles is $N$ (for example $N=3$ in Fig. \ref{fig:models}), PE inputs raw sensor values to the $N$ models, each of which is trained independently. In the original bagging algorithm, multiple subsets are constructed from the original training dataset, and each model is trained using each subset; however, in this section, the same input is used to simplify the explanation of EE. A discussion of the input style is described in Section \ref{sec:input}.

While PE improves the generalization performance, it requires that $N$ models are trained by calculating the loss function $N$ times for each path during training because there are multiple outputs. Further, the inference path differs from the training path, which increases implementation cost. Therefore, when applying an ensemble in deep learning [Fig. \ref{fig:models} (c)], a method that unifies the outputs of each path and uses weighted summation by weight coefficients $\{\lambda_1, \lambda_2, \lambda_3 \}$ is generally adopted (hereinafter, merge ensemble (ME)). ME improves the multioutput nature of PE and is expected to improve generalization performance by ensemble learning with a single-input single-output model. However, it lacks ease of operation because it requires the management of multiple models.

In this study, we propose EE, which improves the need to manage multiple models in ME. The advantages of EE are that it is as tractable and easy to implement as the BL model. $N$-times repeated inputs in the channel direction are independently processed in the EE block (EE-Block) as $N$ groups of channels. EE is a method to realize multiple encoders of ME in a single model. ResNeXt is similar to the transformed each module of ResNet to EE style (with different details such as normalization methods). In other words, EE is a generalization of the concept of ``Cardinality'' in ResNeXt.

As shown in Fig. \ref{fig:models} (d), the model architecture of EE is almost the same as that of BL, with the only differences being that the $N$ inputs to be ensembled are concatenated in the channel direction and the internal module that constitutes E is changed from the general convolutional block to EE-Block. Therefore, EE is easy to handle as a BL model in that it has a single input, output, and model.

\subsection{EE-Block}
The currently proposed modules of various CNNs can be easily converted into EE by simply changing some layers and hyperparameters. First, we list the typical layers used in existing CNNs. For example, in a visual geometry group (VGG) \cite{Simonyan2014}, the encoder comprises a convolutional layer, normalization layer, activation function, and pooling layer. In addition, Xception \cite{Chollet2017} uses depth-wise separable convolution, which divides the convolution layer into a depth and a point-wise convolution layer to reduce the number of parameters. ResNet \cite{He2016} uses a shortcut connection that enables skipping some layers to remove the vanishing gradient issue. Most CNNs are realized by combining these layers.

By modifying each of the abovementioned layer as follows, all CNN modules can be EE style.

\begin{itemize}
    \item Set the hyperparameter of groups as $N$ of the convolution layer to change to group convolution (same for the point-wise convolution layer).
    \item Change normalization layer to group normalization \cite{Wu_2018_ECCV} by setting the group parameter to $N$.
    \item Because the fully-connected layer is equivalent to a one-dimensional convolution layer with a kernel size of 1, change the fully-connected layer to reshape and the group convolution layer.
    \item Multiply output $\bm{z}$ of E by $\frac{1}{N}$.
\end{itemize}
Notably, the activation function, pooling layer, depth-wise convolution layer, and shortcut connection are processed independently in the channel direction; therefore, they do not need to be changed. Using these simple procedures, all CNN architectures can be in the EE style, excluding the concatenation process (described in detail at the end of Section IV-C).

EE is easy to implement using mainstream deep-learning libraries, such as Pytorch and TensorFlow. Because they have a group hyperparameter in the convolution layer, EE can only be implemented by setting the group parameter to $N$ when building the BL model. Further, these libraries have a group normalization layer; thus, only replacing normalization layers with group normalization layers is necessary. Implementing EE requires approximately the same amount of effort as implementing BL. Sample code and implementation procedure are available at \url{https://github.com/t-hasegawa-FU/Easy-Ensemble}.

\subsection{Proof of equality between ME and EE}
We describe the equality between ME and EE [Fig. \ref{fig:models} (c) and (d)] via the proof of equality in each layer type.

\subsubsection{Convolution layer}
\label{sec:conv}
$\bm{X} = [\bm{x}_1, \bm{x}_2, \cdots, \bm{x}_n]$ denotes input tensor. Here each $\bm{x}_i$ is a waveform sensor value in channel $i$, and $n$ is the total number of channels. Three-axis acceleration sensor values (x, y, and z) are commonly used as inputs in HAR. $\bm{Z} = [\bm{z}_1, \bm{z}_2, \cdots, \bm{z}_m]$, which is the output of a single convolution layer in ME, can be calculated as follows:

\begin{equation}
  \label{eq:linear1}
  \bm{z}_i = \sum_{j=1}^{n} \bm{x}_j * \bm{w}_{i,j} + b_i
\end{equation}
where $\bm{w}$ and $b$ denote the filter and bias of the convolution layer, respectively, and $*$ is the cross-correlation operator.

By contrast, the input tensor in EE is defined as $\bm{X'} = [\bm{x}_1, \bm{x}_2, \cdots, \bm{x}_{n \times N}]$ because the input is repeated $N$ times in the channel direction in advance, where $N$ denotes the number of ensembles. $\bm{Z'} = [\bm{z}_1, \bm{z}_2, \cdots, \bm{z}_{m \times N}]$, which is the output of a single group convolution layer in EE, can be calculated as follows:

\begin{equation}
  \bm{z}_i = \left\{
  \begin{array}{ll}
      \sum_{j=1}^{n} \bm{x}_j * \bm{w}_{i,j} + b_i & (i \leq m)\\
      \sum_{j=1}^{n} \bm{x}_{n+j} * \bm{w}_{i,j} + b_i & (m < i \leq 2m) \\
  \vdots & \vdots
  \end{array}
  \right.
\end{equation}

Group convolution, as defined, divides the input channel into $N$ equal groups and performs an operation to convolute independent parameters into each group. Therefore, $\bm{Z'}$ is equivalent to the output of combining $\bm{Z}^{(1)}, \bm{Z}^{(2)}, \bm{Z}^{(3)}$ through $N$ individual convolution layers.

\subsubsection{Normalization layer}
To be EE, all normalization layers need to be group normalization layers. Although ME and EE are not completely equivalent when the normalizations are batch normalizations, assuming that all normalizations in ME are layer normalizations, ME and EE completely equivalent.

Layer normalization in ME normalizes input tensor $\bm{X}$ as follows:

\begin{equation}
    \bm{Z} = \frac{\bm{X} - \bar{\bm{X}}}{\sqrt{(Var(\bm{X}) + \epsilon)}}
\end{equation}
where $Var(\bm{X})$ denotes the variance of $\bm{X}$, $\epsilon$ denotes a small constant (1e-05 in this study). Some implementations may use trainable scaling parameters, but we do not use them in EE.

In group normalization, combined input tensor $\bm{X'} = [\bm{X}^{(1)}, \bm{X}^{(2)}, \cdots, \bm{X}^{(N)}]$ (the shape of each $\bm{X}^{(i)}$ is as same as $\bm{X}$) is normalized to output $\bm{Z'} = [\bm{Z}^{(1)}, \bm{Z}^{(2)}, \cdots, \bm{Z}^{(N)}]$ as follows:

\begin{equation}
\bm{Z}^{(i)} = \frac{\bm{X}^{(i)} - \bar{\bm{X}}^{(i)}}{\sqrt{(Var(\bm{X}^{(i)}) + \epsilon)}}
\end{equation}

Group normalization, as defined, divides the input channel into $N$ equal groups and applies layer normalization to each group. Therefore, $\bm{Z'}$ is equivalent to the output of each layer normalization combined $N$ times.

\subsubsection{Fully-connected layer}
$\bm{v} = [v_0, v_1, \cdots, v_n] \in \mathbb{R}^{n}$ denotes the input to a fully-connected layer. According to Equation (\ref{eq:linear1}), transform $\bm{v}$ into $\bm{V} \in \mathbb{R}^{n \times 1}$ and input it into convolution layer with kernel size of 1, and each element $\bm{z}_i$ of output $\bm{Z} \in \mathbb{R}^{m \times 1}$ is calculated as follows:

\begin{equation}
    \bm{z}_i = \sum_{j=1}^{n} \bm{v}_j * \bm{w}_{i,j} + b_i
\end{equation}
where each $\bm{z}_i, \bm{v}_j, \bm{w}_{i,j}$ is substantially a scalar. Thus, the element is as follows:
\begin{equation}
    \label{eq:linear2}
    z_i = \sum_{j=1}^{n} v_j \cdot w_{i,j} + b_i
\end{equation}
As Equation (\ref{eq:linear2}) is equivalent to the definition of a fully-connected layer, the fully-connected layer can be represented by a single convolution layer by reshaping input $\bm{v}$ to $\bm{V} \in \mathbb{R}^{n \times 1}$ and output $\bm{Z}$ to $\bm{z} \in \mathbb{R}^{m}$. Thus, as shown in Subsection \ref{sec:conv}, the fully-connected layer can be EE style by setting groups in the transformed convolution layer.

\subsubsection{Pooling and output layers}
The pooling layer applied between each layer is processed channel-independently and does not affect EE. The encoder output $\bm{z}$ in Fig. \ref{fig:models} (d) is a feature vector of length $m \times N$ obtained by applying global average pooling (GAP) to the final output $\bm{Z}$ of EE-Blocks. The GAP also calculates the average for each channel and does not affect EE.

Finally, using the feature vector $\bm{z}$, weight hyperparameter $\lambda = \frac{1}{N}$, and a single fully-connected layer, EE outputs $\bm{p}$ (applying the softmax function to $\bm{p}$ implies a class probability distribution) as follows:

\begin{eqnarray}
    \label{eq:linear}
    \bm{p} & = & \frac{1}{N} \sum_{i=1}^{m \times N} z_i w_i + b  \nonumber \\
           & = & \frac{1}{N} (\sum_{i=1}^{m} z_i w_i + \cdots + \sum_{i=(N-1)m+1}^{m \times N} z_i w_i) + b
\end{eqnarray}
Here, each member expanded by Equation (\ref{eq:linear}) is equivalent to the output of each path in Fig. \ref{fig:models} (c). Denoting $\lambda=\frac{1}{N}$, Figs. \ref{fig:models} (c) and (d) are equivalent. By vectorizing $\lambda$, it is also possible to set individual weight coefficients for each group, as in Chen et al.'s GE \cite{chen2020group}. However, the verification results did not show any improvement in accuracy; therefore, the same value of $\lambda=\frac{1}{N}$ was used in this study.

EE is a single-model architecture similar to BL but requires multiplying $\lambda$ by the encoder's output $\bm{z}$. In general, a classification model is optimized based on categorical cross-entropy loss using output $\bm{p}$ normalized by the softmax function, as follows:

\begin{equation}
    \label{eq:soft}
    Softmax(p_i, \bm{p}) = \frac{\exp(\frac{p_i}{T})}{\sum_{p_j \in \bm{p}}\exp(\frac{p_j}{T})}
\end{equation}
where $\bm{p}$ denotes the output vector, $p_i$ denotes each element of $\bm{p}$, and is the normalization target. Because EE is equivalent to ME as mentioned above, calculating the predicted label $p_i$ for each path in ME without multiplying by $\lambda$ would result in $N$ times the scale of $p_i$ compared with BL. In Equation (\ref{eq:soft}), $T$ is a temperature parameter that works as the general softmax function at $T=1$, emphasizing higher and lower values when $T<1$ and $T>1$ respectively. The predicted label $p_i$ multiplied by $N$ is the same as the temperature parameter $T=\frac{1}{N}$ in the softmax function that emphasizes high output values. Experimentally, this tends to reduce the estimation accuracy more than with $T=1$. Therefore, in this study, $T=N$, i.e., the output is multiplied by $\frac{1}{N}$.

\section{Experiments}
\subsection{Experimental settings}
We performed experiments to evaluate the effectiveness of our method --EE-- using public benchmark datasets for sensor-based HAR. Although we performed experiments under various conditions, we adopt the experimental settings we describe in this section as the default unless otherwise stated.

The dataset is HASC\cite{Kawaguchi2011}, which comprises accelerometer data labeled with six basic activities (stop, walk, jog, skip, go upstairs, and go downstairs). We extracted data with a sampling frequency of 100 Hz from BasicActivity from 2011 to 2013 in HASC. As preprocessing, we removed the data two seconds before and after each measurement file and extracted 256 sample windows with stride = 256 using the sliding-window method. From the preprocessed data, 10 and 50 randomly selected subjects were classified as a training dataset ($D_{train}$) and testing dataset ($D_{test}$), respectively.

To validate the impact of ensemble learning, the encoder is modeled as a shallow VGG with eight layers. The encoder outputs a feature vector $\bm{z}$ using GAP and predicts activity labels using one fully-connected layer. Layer normalization was used as the normalization method. This model is denoted as BL and is ensembled with $N=4$ according to Fig. \ref{fig:models} (b)--(d) for comparison. \update{In this study, for ME, $\{\lambda_i = \frac{1}{N} | i=1,2,.... .N \}$, which is the same as multiplying $\frac{1}{N}$ to $\bm{z}$ in EE. EE and ME can adopt more sophisticated strategies as an aggregation algorithm, such as soft combiners \cite{ACETO2018131} and group wagging \cite{chen2020group}. Since EE and ME are equivalent, adapting such strategy can improve the generalization performance also of EE. These strategies can be implemented in EE by multiplying $\bm{\lambda}$ vector of length equal to the number of channels in $\bm{z}$ rather than multiplying $\frac{1}{N}$ to $\bm{z}$.}

For all models, the optimization method was Adam with 1e-3 of a learning rate, and 500 epochs were trained with a batch size of 1,000. Rotation, which randomly applies channel shuffling and positive/negative inversion, was adopted for data augmentation because IoT devices are sometimes equipped with random orientations. The evaluation index was accuracy, and the results of multiple trials with different random seed numbers were discussed without changing the subjects in each dataset. All experiments were implemented using python 3.7 and pytorch 1.7 in a cuda environment on computers equipped with NVIDIA GPUs.

\subsection{Evaluation on effectiveness and equivalence}
\begin{figure}[b]
    \begin{center}
    \includegraphics[width=5.5cm]{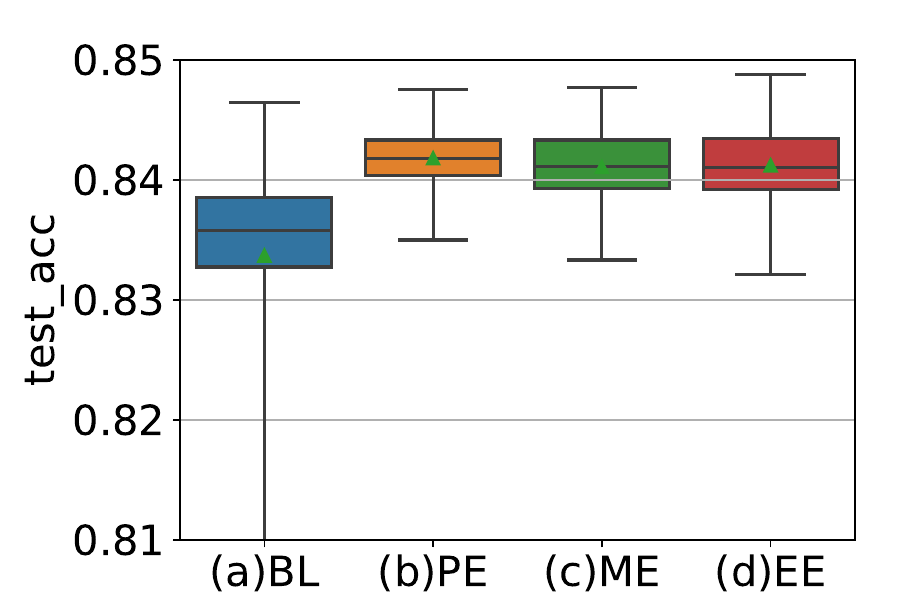}
    \caption{Ensemble method (VGG architecture) versus HASC Accuracy. The number of filters in the first convolutional layer is 16.}
    \label{fig:exp1_basic}
    \end{center}
\end{figure}

Fig. \ref{fig:exp1_basic} shows the results of 100 trials of the HASC dataset for each ensemble model, where BL is the aggregate results of 100 trials of the four models generated during PE training (total 400 trials). Notably some results are outside the range of the figure because of some divergent results.

\begin{figure*}[tb]
    \begin{center}
    \includegraphics[width=18.0cm]{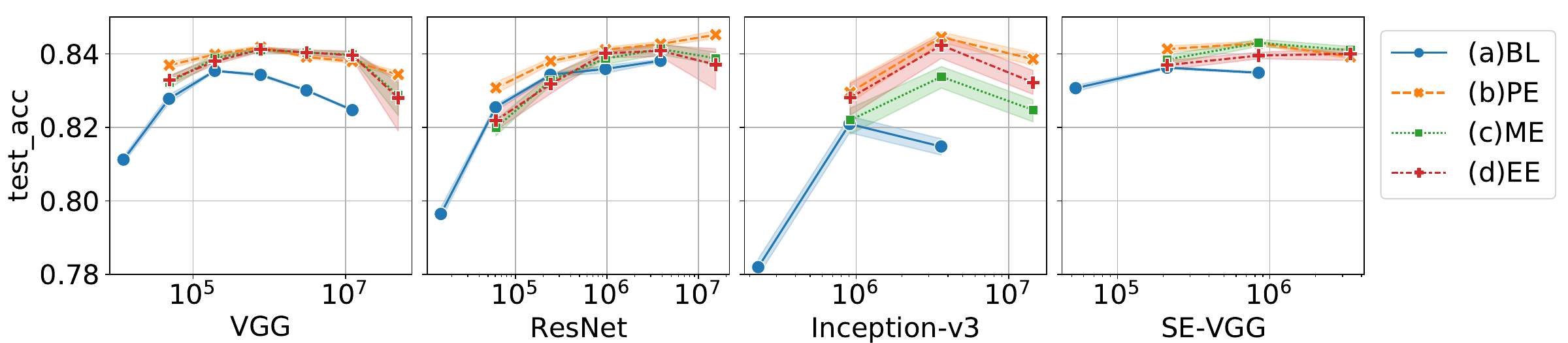}
    \caption{Model size versus HASC accuracy for each model.}
    \label{fig:exp2_params}
    \end{center}
\end{figure*}

From the figure, the ensemble models generally have a higher estimation accuracy than the single model BL, and ME and EE are generally equivalent in terms of estimation accuracy.

\subsection{Effects of the model architecture and number of parameters}
From Fig. \ref{fig:exp1_basic}, only BL has few trainable parameters under a unified encoder condition. In addition, the model architectures were limited to VGG. We evaluated the effectiveness of various model architectures considering the model size. We implemented ResNet \cite{He2016}, Inception-v3 \cite{Szegedy2016}, and SENet \cite{Hu2018} (on VGG), which are commonly used as baseline models in image recognition. Each model was modified to EE and other ensemble styles. Fig. \ref{fig:exp2_params} shows the change in estimation accuracy with an increase in model size. The results of each model, excluding VGG, were summarized from 50 trials, whereas that of VGG was from 100 trials. The horizontal and vertical axes of Fig. \ref{fig:exp2_params}, respectively, indicate the number of parameters and estimation accuracy of the test data. Each model was implemented by multiplying the number of filters in each convolutional layer by $x$. Notably, we excluded few results that did not converge ($accuracy(D_{test}) \leq 0.5$) and showed the mean and 95\% confidence interval in the figure. These results represent the $N=4$ ensembles, and doubling the number of filters in each layer of the BL model is approximately equivalent to the number of parameters in the $N=4$ ensemble model.

The results of VGG indicated the following four points.
\begin{itemize}
    \item Under the same model size, ensemble methods outperform BL.
    \item The accuracy difference among the ensemble methods is not significant; it is approximately $PE \geq  ME \simeq EE$.
    \item Because of the limited size of the training dataset, if the model size becomes too large, the estimation accuracy decreases.
    \item The changes in the model size and accuracy are generally similar for BL and ensemble methods.
\end{itemize}
\begin{figure}[b]
    \begin{center}
    \includegraphics[width=6.5cm]{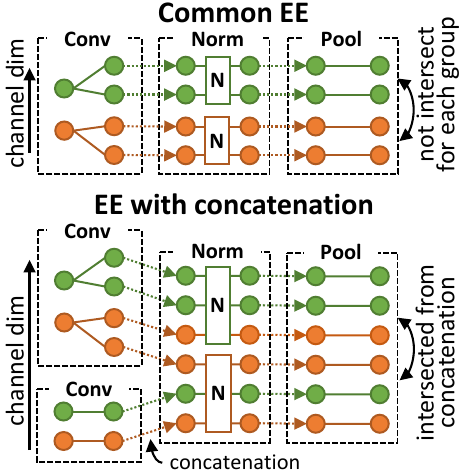}
    \caption{Intersected channels from concatenation process.}
    \label{fig:channelshuffle}
    \end{center}
\end{figure}

Focusing on other model architectures, the results showed a similar trend to that of VGG, indicating that EE worked well for various model architectures. However, the results also suggested that, in ResNet, EE was sometimes inferior to BL if an appropriate model size was not set. In addition, in Inception-v3, EE outperformed ME, whereas PE slightly outperformed EE in terms of accuracy. As mentioned in Section III-B, all CNN architectures can be in the EE style, excluding the concatenation process. However, because Inception-v3 has concatenation, the EE-styled Inception-v3 is distinct from the ME-styled Inception-v3.

Fig. \ref{fig:channelshuffle} illustrates the difference between common EE and EE with concatenation. The concept of EE is to create groups in the channel direction and execute process for each layer on a group basis (upper part of Fig. \ref{fig:channelshuffle}); however, channels of each group are intersected by the concatenation process (lower part of Fig. \ref{fig:channelshuffle}). It is possible to convert a model including the concatenation process to EE style by reordering the channel before concatenating. However, since Inception-v3 results in Fig. \ref{fig:exp2_params} indicate improved estimation accuracy, we conclude that it is unnecessary to reproduce the EE-styled model including concatenation. Instead, we investigated why the accuracy improved. The reason is that the concatenation process worked like channel shuffle proposed in ShuffleNet \cite{Zhang2018_shuffle}. Group convolution contributes to reducing the number of parameters, but it limits the input information in a lightweight model \cite{Wang_2019_CVPR}. Channel shuffle solves this problem by reshaping and transposing tensors to swap channels across groups. The EE model is an ensemble model, but it functions as a channel shuffle during concatenation, which is believed to have contributed to the accuracy improvement in Inception-v3.
\begin{table}[b]
  \newcolumntype{A}{>{\centering\arraybackslash}p{2em}}
  \newcolumntype{B}{>{\centering\arraybackslash}p{6em}}
  \newcolumntype{C}{>{\centering\arraybackslash}p{4em}}
  \caption{Detailed information of datasets}
  \label{table:datasets}
  \hbox to\hsize{\hfil
  \begin{tabular}{c|AA|BC} \hline \hline
      \multirow{2}{*}{Dataset} &	\multicolumn{2}{c|}{\# of subj. in} &	\multicolumn{2}{c}{Shape of} \\ 
      &	train &	test &	input &	output \\ \hline
     HASC \cite{Kawaguchi2011} &	10 &	50 &	(?, 3, 256) &	(?, 6) \\ 
     UCI Smartphone \cite{Anguita2013} &	21 &	9 &	(?, 9, 128) &	(?, 6) \\ 
     WISDM \cite{Kwapisz2011} &	26 &	11 &	(?, 3, 256) &	(?, 6) \\ 
     UniMiB SHAR \cite{Daniela2017} &	21 &	9 &	(?, 3, 151) &	(?, 17) \\ 
     PAMAP2 \cite{Reiss2012a} &	7 &	2 &	(?, 42, 256) &	(?, 12) \\ \hline
  \end{tabular}\hfil}
\end{table}
\begin{figure*}[tb]
    \begin{center}
    \includegraphics[width=18.0cm]{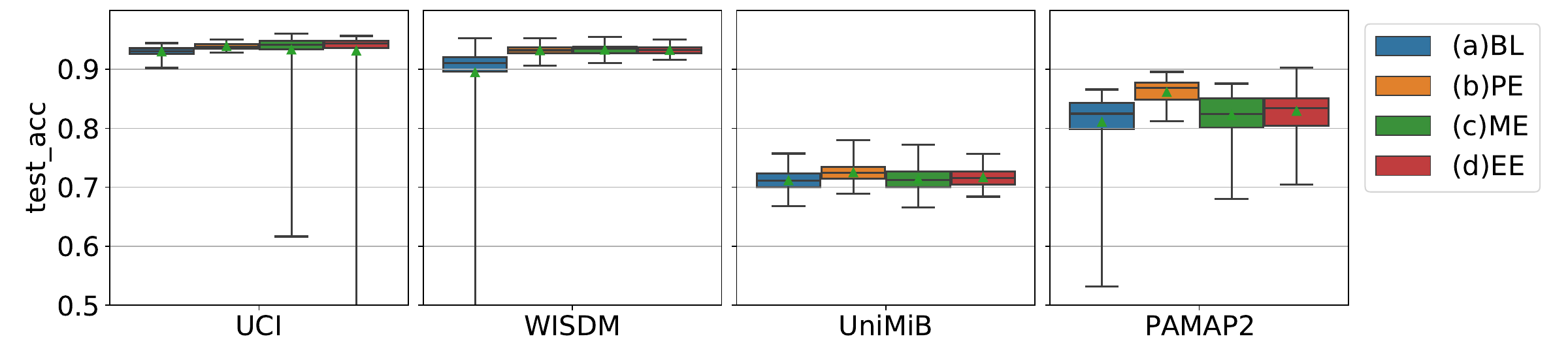}
    \caption{HAR accuracies for each public dataset (VGG architecture).}
    \label{fig:exp3_datasets}
    \end{center}
\end{figure*}

\subsection{Robustness of target domains}
In this section, we evaluate the robustness of EE using several public datasets for sensor-based HAR. The additional datasets used were the UCI Smartphone Dataset \cite{Anguita2013}, WISDM \cite{Kwapisz2011}, UniMiB SHAR \cite{Daniela2017}, and PAMAP2 \cite{Reiss2012a}. Table \ref{table:datasets} lists the details of the datasets, including HASC. PAMAP2 has 52 dimensions of sensor data, but we excluded orientation (12 dimensions), which is officially invalid. For the convenience of the process to change the input method described below, we rearranged the order of dimensions and copied only the temperature from one to three dimensions.
\begin{figure}[t]
    \begin{center}
    \includegraphics[width=5.5cm]{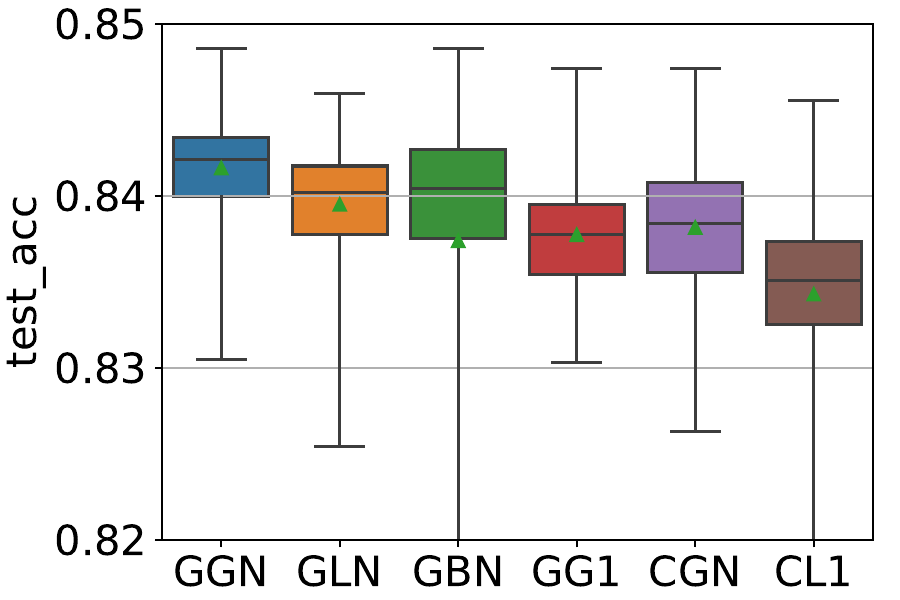}
    \caption{Ablation study (VGG architecture and HASC accuracy). The first letter denotes convolution type (G: group convolution and C: conventional convolution). The second letter denotes normalization type (G: group normalization, L: layer normalization, and B: batch normalization). The third letter denotes the weights of each ensemble path (N: divided by \# of ensembles $N$ and 1: divided by 1). CL1 is the same as (a) BL.}
    \label{fig:exp4_ablation}
    \end{center}
\end{figure}

Fig. \ref{fig:exp3_datasets} shows the results of a 50-trial experiment; in all datasets, PE outperformed BL, whereas ME and EE were generally equivalent. Moreover, ME and EE were equivalent PE in UCI and WISDM, whereas PE outperformed ME and EE in UniMiB and PAMAP2. One of the characteristics of UniMiB and PAMAP2 is numerous classes, indicating that the effectiveness of ensembles with a single-input single-output model using ME and EE depends on the dataset.

%\subsection{Considerations for the characteristics of EE}
\subsection{Ablation Study}
We conducted an ablation study to investigate the factors that improved the accuracy of EE. Fig. \ref{fig:exp4_ablation} shows 100-trial results. The first letter denotes convolution type (G: group convolution and C: conventional convolution). The second letter denotes normalization type (G: group normalization, L: layer normalization, and B: batch normalization). The third letter denotes the weights of each ensemble path (N: divided by \# of ensembles $N$ and 1: divided by 1). As such, GGN is the same as (d) EE and CL1 is the same as (a) BL.

We consider Fig. \ref{fig:exp4_ablation} from the viewpoint of ablation study. A 0.5\% decrease in accuracy was due to group convolution becoming a conventional convolution (GGN versus CGN) and the lack of $\frac{1}{N}$ when merging ensemble results (GGN versus GG1). In addition, replacing group normalization with layer normalization resulted in a slight decrease in accuracy. The best result was EE (GGN), which incorporated the three factors (group convolution, group normalization, and multiplying $\frac{1}{N}$). The accuracy decreased when group normalization was substituted with batch normalization. 

\section{Input variationer}
In this section, we propose and evaluate the effectiveness of the input variationer shown in Fig. \ref{fig:abstract} as a method to add variety to the input given to each group in EE. In ensemble learning, such as bagging, different subsets are assigned to each classifier to obtain diversity. However, so far, we have used the function $R_N()$ to repeat the original input ($\bm{X}\in \mathbb{R}^{b \times c \times w}$) $N$ times in the channel direction as $R_N(\bm{X}) \in \mathbb{R}^{b \times cN \times w}$ to generate the input into EE, where $b$, $c$, and $w$ denote the batch size, number of channels, and window size respectively. In other words, the accuracy of EE can be further improved by diversifying the input because the diversity of EE's input so far has been given only by the variation in the initial values of network parameters.

\subsection{Types of input variationer}
\label{sec:input}
The ensemble method using data augmentation (DA) (hereinafter, $A_N()$, using the number of ensembles $N$) has been used in several existing studies. For example, in a previous study \cite{Hasegawa2021}, we proposed a method to ensemble DAs, such as Mixup \cite{Zhang2018} and RICAP \cite{Takahashi2018}, into our original DA method, OctaveMix. In the previous study, DA was fixed for each path; however, in this study, we decided to apply multiple DAs randomly in each path, based on the idea of RandAugment \cite{Ekin2020}. To formulate, $A_N(\bm{X}, \bm{t}) = cat(t_{i_0}(\bm{X}), t_{i_1}(\bm{X}), \cdots)$, where $\bm{t} = \{t_{i} | i=0,1, \cdots, N \}$ is a DA set of type i; $i_0, i_1, \cdots $ is the index number determined by random numbers without overlap; cat() is the process of combining each tensor in the channel direction.

Another typical method in HAR is a method of training and ensembling the data observed by each sensor modality with independent models (hereinafter, \update{modality ensemble (Mod)}). Notably, the UCI Smartphone Dataset \cite{Anguita2013} combines total acceleration (3ch), estimated body acceleration (3ch), and angular velocity (3ch) data in the channel direction into nine channels of data. In Mod, when inputting data into EE, no additional processing is required on the DataLoader side, but an appropriate number of groups $N$ must be set on the EE side. 
\update{As related work, Aceto et al. proposed a multi-modal deep ensemble learning, which used merged feature vectors extracted from multiple independent deep learning models for each modality \cite{MIMETIC}. Sena et al. also proposed an ensemble deep learning method combining Mod with CNN ensemble with various kernel sizes \cite{SENA2021226}. EE can easily implement Mod only when each architecture is the same.}

\begin{figure}[t]
    \begin{center}
    \includegraphics[width=8.5cm]{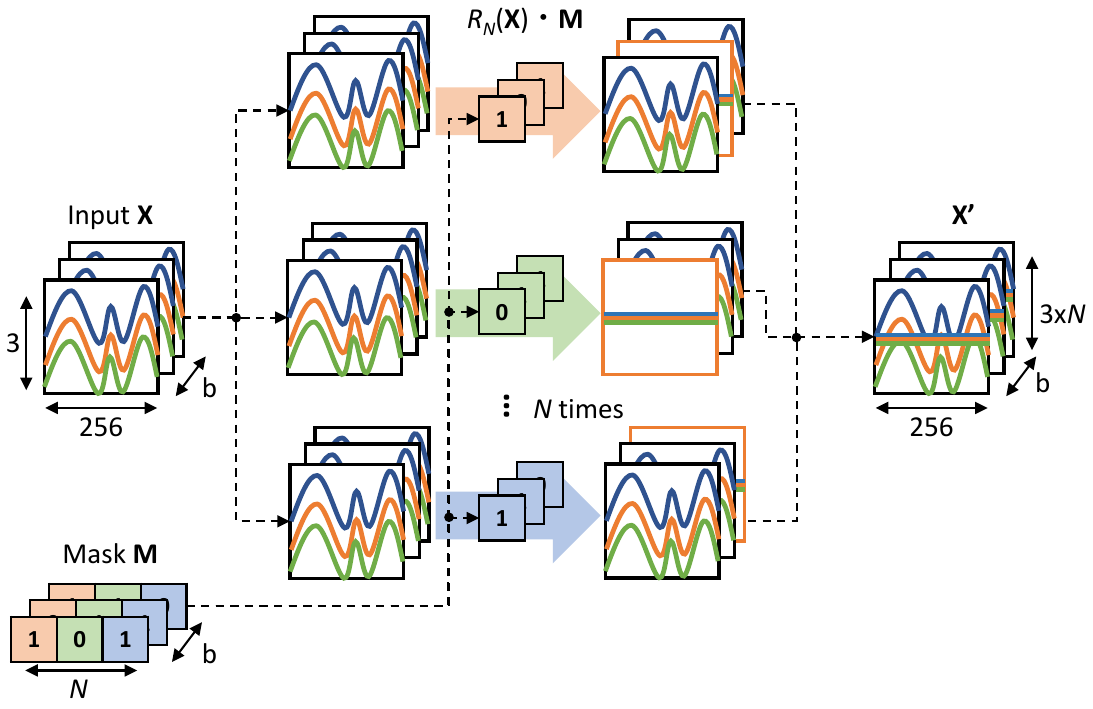}
    \caption{Input masking to make input groups.}
    \label{fig:inputmask}
    \end{center}
\end{figure}

As a new type of input variationer, we propose input masking, which incorporates the idea of bootstrap sampling. In ensemble methods, such as ME and EE, which combine the output parts, it is necessary to handle the input $\bm{X}$ corresponding to the same correct label $\bm{y}$ among the paths of the ensemble to conveniently compute the loss at once. Therefore, in input masking, preparing a mask tensor $\bm{M} \in \mathbb{B}^{b \times N}$ ($\mathbb{B}=\{0, 1\}$) in the DataLoader, masked output $\bm{X}_{M} = Mask(\bm{X}, \bm{M}) = R_N(\bm{X}) \cdot \bm{M}$ is calculated (Fig. \ref{fig:inputmask}). Input masking is a method to construct a pseudosubset of $\bm{X}_{M}$ by masking certain instances of $\bm{X}$ with zeros. The output is $\bm{X}_{M} \in \mathbb{R}^{b \times cN \times w}$. Any method for generating $\bm{M}$ is acceptable; in this study, we assign a random mask as in $N$-fold cross-validation and design the mask such that $N-1$ groups are always valid in each instance. Accordingly, we changed $\frac{1}{N}$ multiplied by the feature map to $\frac{1}{N-1}$.

\subsection{Evaluating input variationer for a single modality dataset}
\begin{figure}[tb]
    \begin{center}
    \includegraphics[width=6.5cm]{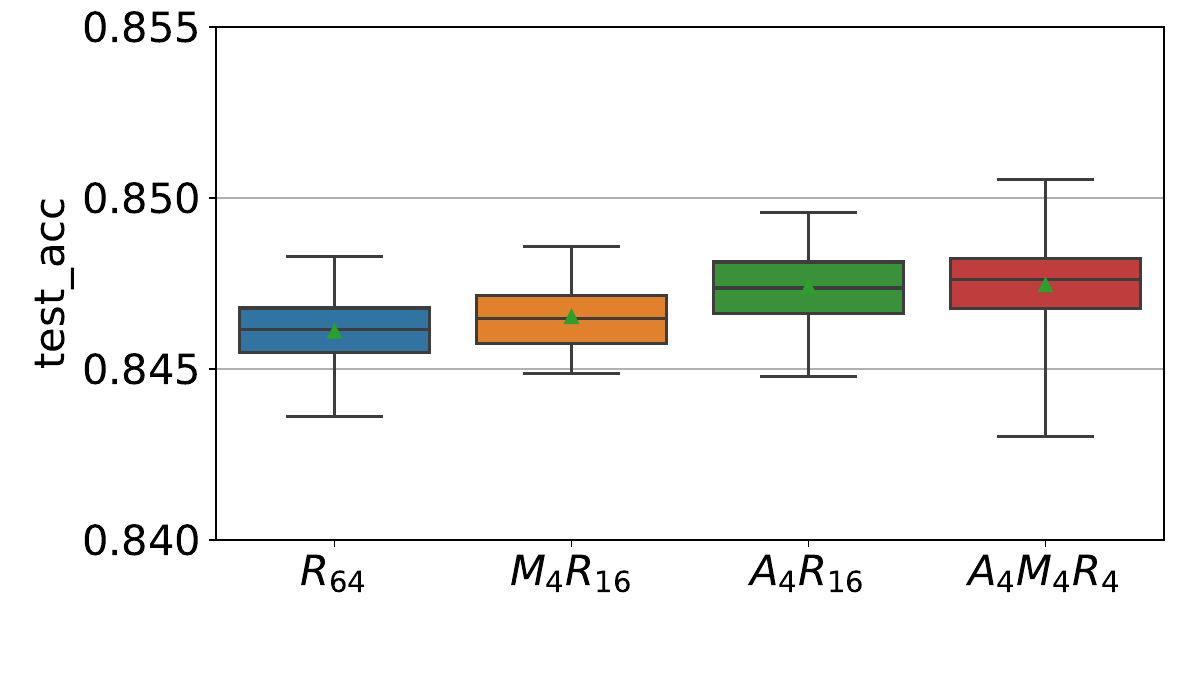}
    \caption{HASC accuracies for each input variation (VGG architecture).}
    \label{fig:exp6_inputs}
    \end{center}
\end{figure}

First, we evaluated the effectiveness of the input variationer on data measured using a single sensor modality, such as the HASC dataset. The number of ensembles was unified at $N=64$, and combinations of repeat ($R_N$), augmentation ($A_N$), and input masking ($M_N$) were evaluated. Fig. \ref{fig:exp6_inputs} shows the results of 50 trials for each method, where $R_{64}$ denotes the method applying EE with $N=64$, $M_4R_{16}$ denotes the method applying input masking with $N=4$ to the repeated input $R_{64}$, $A_4R_{16}$ is the method applying four data augmentations to $R_4$ and repeating them $N=16$ times, and $A_4M_4R_4$ is a method in which $N=4$ times repeats the augmentation result $A_4$, applies input masking of $N=4$, and repeating $N=4$ times. In this study, we adopted four types of DAs: two types of jittering with different magnitudes, scaling of waveform amplitude, and shift in amplitude direction. Notably, rotation was applied in all cases.

As shown in Fig. \ref{fig:exp6_inputs}, input masking ($M_4R_{16}$) is slightly better than repetition ($R_{64}$), but the augmentation ($A_4R_{16}$) and combination of augmentation and input masking ($A_4M_4R_4$) add diversity to the input and improve the HAR accuracy by a median of 0.15\%.

\subsection{Evaluation of input variationer for multimodality datasets}
\begin{figure}[tb]
    \begin{center}
    \includegraphics[width=8.5cm]{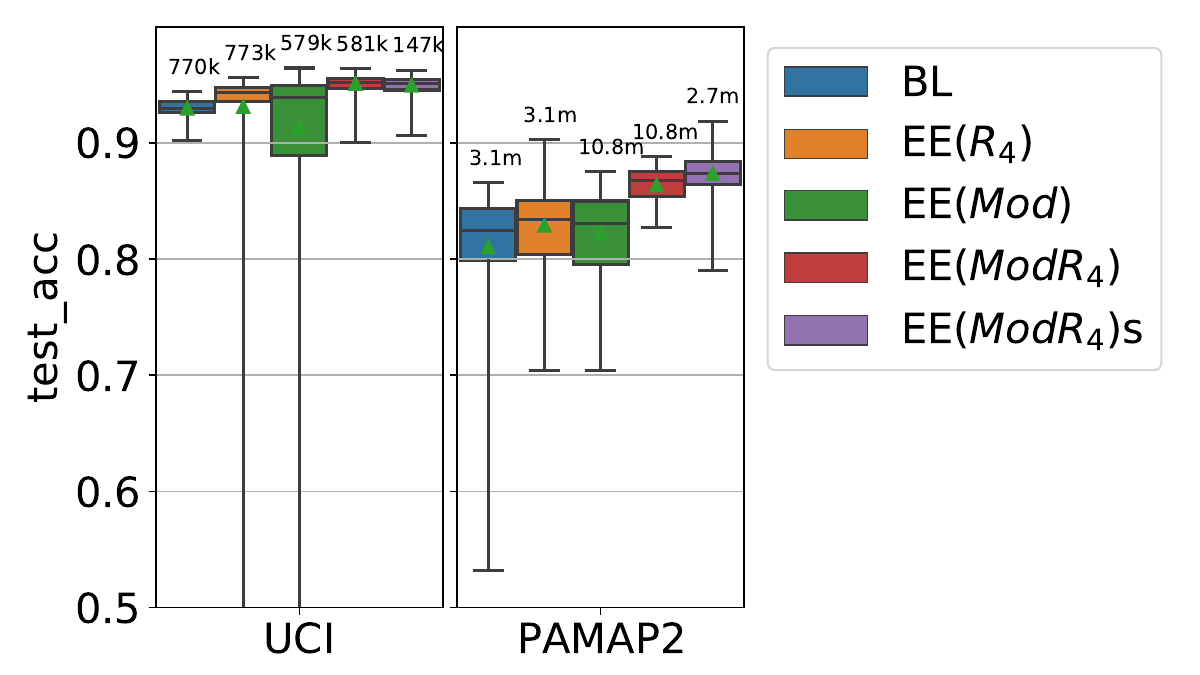}
    \caption{HAR accuracies by modality ensembles (VGG architecture).}
    \label{fig:exp7_modality}
    \end{center}
\end{figure}

Next, we evaluated the effect of the input variationer on data measured with multiple modalities, such as UCI and PAMAP2. Fig. \ref{fig:exp7_modality} shows 50-trial retults using a combination of repeat ($R_N$) and modality ensemble ($Mod$). EE($R_4$) is EE with repeat for $N=4$, $Mod$ is the modality ensemble, and $ModR_4$ is the case where repeat is used with the modality ensemble. Because the number of parameters in $ModR_4$ is too large, the number of parameters is adjusted by the filter size in EE($ModR_4$)s. The number of parameters for each model is shown in each box plot.

From the figure, the best accuracy was obtained using repeat with modality ensemble for both UCI and PAMAP2. In particular, for PAMAP2, the best accuracy was achieved with lightweight EE ($ModR4$)s, which was lighter than BL and approximately 5\% more accurate.

\section{Discussion}
\subsection{Qualitative analysis}
\begin{table}[b]
    \newcolumntype{A}{>{\centering\arraybackslash}p{12em}}
    \caption{Qualitative evaluation of ensemble methods.}
    \label{table:qualitative}
    \scalebox{1.0}{
    \hbox to\hsize{\hfil
    \begin{tabular}{c|cccc} \hline \hline
        &	(a) BL &	(b) PE &	(c) ME &	(d) EE \\ \hline
        Ease of implementation &	Good &	Good &	Poor &	Fair \\ 
        Ease of use &	Good &	Poor &	Good &	Good \\ 
        Model management &	Good &	Poor &	Fair &	Good \\ 
        Multiple architectures &	Poor &	Good &	Good &	Poor \\ 
        Stepwise ensemble &	Poor &	Fair &	Fair &	Good \\
        Accuracy &	Poor &	Good &	Good &	Good \\ \hline
    \end{tabular}\hfil}
    }
\end{table}
We qualitatively discuss the differences among the ensemble methods. The differences among the four methods discussed in this study are summarized in Table \ref{table:qualitative}. When using BL as the base for implementation, PE is easy to implement because it requires only multiple models, whereas ME requires the input and output parts to be branched and combined. In EE, the input must be coupled in the channel direction, but the output is the same as in BL. In terms of ease of use during training and inference, ME and EE can be used similarly as BL, but PE requires multiple models to be trained and different processes between the training and inference. In terms of model management, such as weight storage, BL and EE only need to manage a single model, whereas PE and ME need to manage multiple models.

The limitation of EE is that various model architectures cannot be ensembled simultaneously: Mukherjee et al. \cite{Mukherjee2020} ensembled CNN-Net, Encoded-Net, and CNN-LSTM feature extractors, and Dua et al. \cite{Dua2021} ensembled multiple CNN-GRU-based feature extractors with different kernel sizes. Although ensembling feature extractors with various architectures vary feature representations and improve generalization performance, it is time-consuming to implement various architectures. Moreover, although EE can only ensemble the same architectures, it has the advantage of having the number of ensembles $N$ as a hyperparameter, making it easy to implement extended models as a stepwise variation of $N$ among blocks (as described in the next section).

Another advantage of EE is that it allows a single DataLoader and model to divide the roles of the ensemble diversity acquisition. While PE and ME (where the input is not repeated) require the manipulation of multiple DataLoaders, EE requires only a single DataLoader that performs data shaping based on the rule of channel-wise merging. Therefore, users only need to develop a DataLoader for their own dataset and incorporate the input variationer functionality into the DataLoader, which can be reused by simply changing the hyperparameters of the EE model (not requiring the re-development of the model).

In summary, EE is a method that is as easy to use as BL but can benefit from improved accuracy due to its ensemble nature. Compared with PE and ME, EE is easier to implement, and it is easier to implement extensions such as stepwise ensembling.

\subsection{Stepwise ensemble}
\begin{figure}[b]
    \begin{center}
    \includegraphics[width=5.5cm]{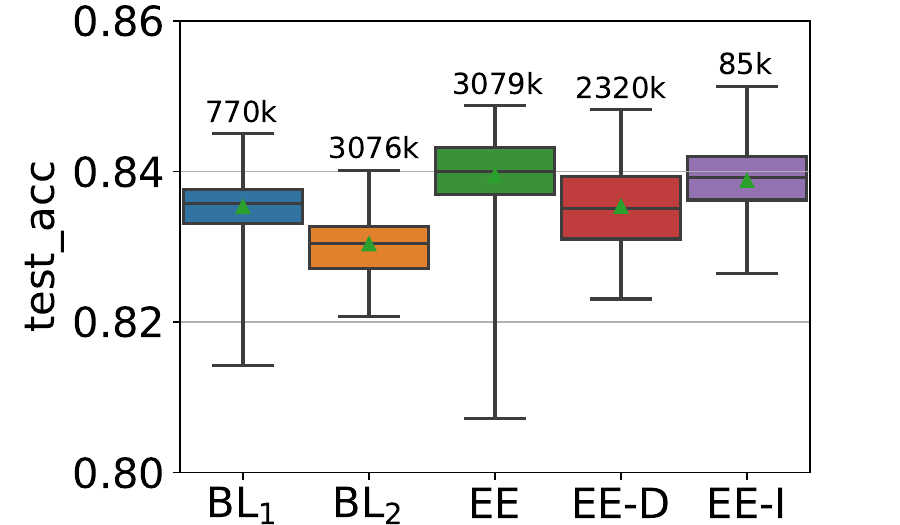}
    \caption{The effect of stepwise ensemble model (HASC acc. and VGG architecture).}
    \label{fig:exp8_stepby}
    \end{center}
\end{figure}

For the five blocks in the VGG architecture, we set all $N=4$ for EE, $\bm{N} = \{4, 4. 2, 2, 1\}$ for the EE-D, and $\bm{N} = \{1, 2, 4, 8, 16\}$ for the EE-I. Fig. \ref{fig:exp8_stepby} shows the results of 100 trials of ensemble learning, wherein $N$ is varied in steps between blocks. While EE had the best estimation accuracy, EE-I achieved similar accuracy with a model that reduced the number of parameters by approximately 97\%. Stepwise ensembling was also discussed in Chen et al.'s study \cite{chen2020group}; depending on hyperparameter tuning, this method is expected to provide ensembling benefits with fewer parameters.

\subsection{Channel shuffle in VGG}
\begin{figure}[t]
    \begin{center}
    \includegraphics[width=8.5cm]{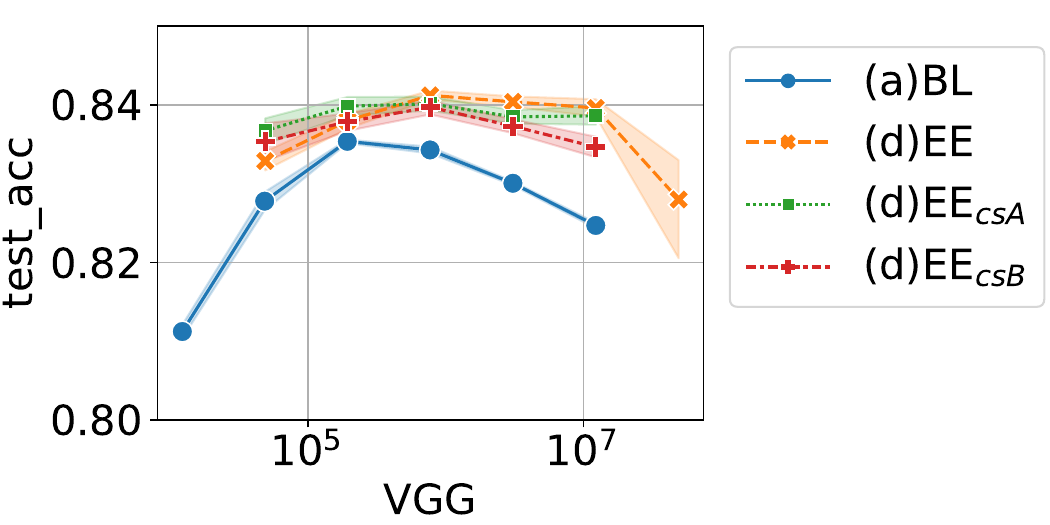}
    \caption{The effect of channel shuffling (HASC acc. and VGG architecture).}
    \label{fig:channelshuffle2}
    \end{center}
\end{figure}

As we mentioned in Section IV-C, channel shuffle can improve the estimation accuracy in EE. The channel shuffle layer can be inserted into EE, which is composed of group convolutions. Fig. \ref{fig:channelshuffle2} depicts the results of 50 trials of EEs for each model. (a)BL and (d)EE are the same as those in Fig. \ref{fig:exp2_params}. (d)EE$_{csA}$ and (d)EE$_{csB}$ are inserted channel shuffle layer at the end of each EE block and normalization after convolution layer, respectively. Notably, (d)EE$_{csA}$ worked well with fewer filters. This trend is similar to that of Inception-v3 (Fig. \ref{fig:exp2_params}). Although Inception-v3 has deeper architecture than the others, its number of filters for each layer was few. Therefore, the accuracy improvement between EE and ME in Fig. \ref{fig:exp2_params} is most likely due to the channel shuffle in Inception-v3.

\subsection{Model scale up}
\subsubsection{Model size and number of ensembles}
\begin{figure}[t]
    \begin{center}
    \includegraphics[width=8.5cm]{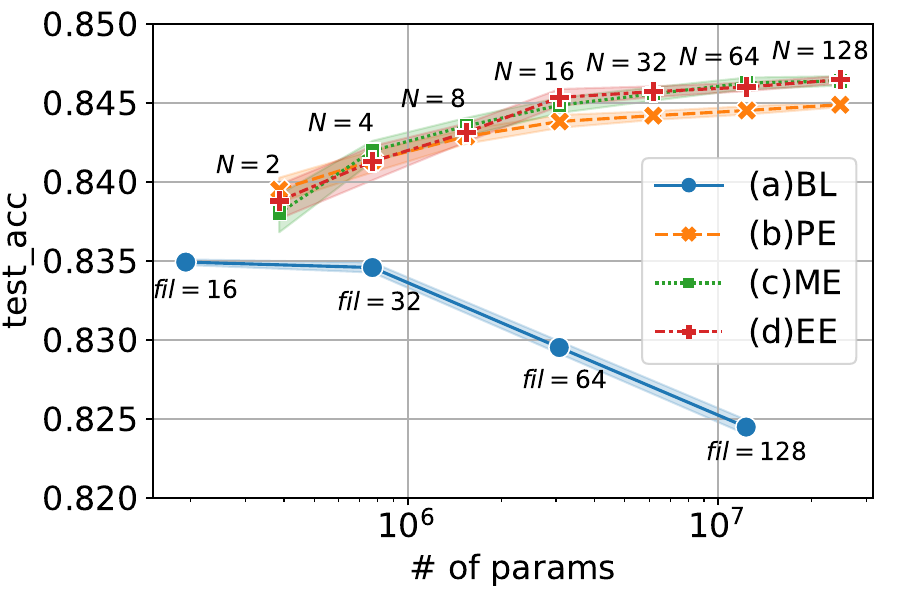}
    \caption{The effect of scale up (HASC acc. and VGG architecture). Ensemble models are scaled up by increasing the number of ensembles ($N$). The number of filters in the first convolutional layer is 16 in ensemble models.}
    \label{fig:exp5_ensemble}
    \end{center}
\end{figure}

Fig. \ref{fig:exp5_ensemble} shows the accuracy of the HASC (50 trials) when the EE of VGG is scaled up by the number of ensembles $N$. As a comparison, the accuracy of BL scaled up by filter size is shown.

From the figure, the estimation accuracy of the scale-up by filter size reaches its peak around $fil=32$ in BL, whereas the accuracy of the scale-up by the number of ensembles $N$ continues to improve as $N$ increases in PE, ME, and EE. In Fig. \ref{fig:exp2_params}, EE also tried to scale-up by filter size and showed the same tendency of reaching its peak as BL.

Focusing on the accuracy of PE and EE, the accuracy of PE was greater than or equal to that of EE up to approximately $N\leq8$, whereas the accuracy of EE exceeded that of PE for $N>8$. Unlike EE, which shares a single loss, it was expected that PE, wherein each path calculates the loss independently, would have higher accuracy; however, when $N$ is large, the results are different. For PE, each model is trained independently; meanwhile, for EE and ME, each model is treated as a single model, which may yeild to the acquisition of complementary feature representations. This result demonstrates the usefulness of the proposed method.

\subsubsection{The number of filters and ensembles}
\begin{figure}[t]
    \begin{center}
    \includegraphics[width=8.5cm]{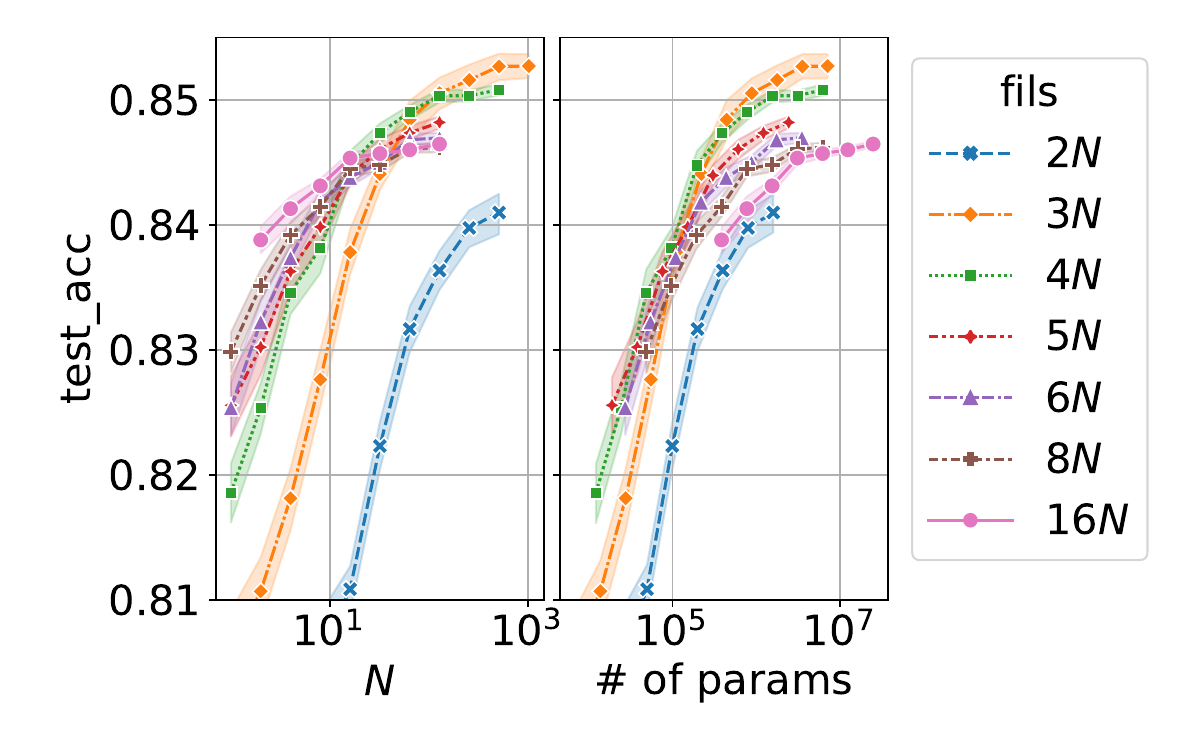}
    \caption{The effect of the number of filters in the first convolutional layer (HASC acc. and VGG architecture). Ensemble models are scaled up by increasing the number of ensembles ($N$).}
    \label{fig:exp10_ens_fils}
    \end{center}
\end{figure}

To compare the effect of the number of filters, Fig. \ref{fig:exp10_ens_fils} shows the accuracy of the HASC (50 trials) when the EE of VGG is scaled up by the number of ensembles $N$. Both the left and right sides indicate the same graph but they have different x-axis. Each line denotes the results of the EE model with $x \times N$ filters in the first convolutional layer. This is the same as the $N$-ensembled BL model with $x$ filters in the first convolutional layer.

From Fig. \ref{fig:exp10_ens_fils}, we found that models with a fewer number of filters outperformed those with a larger number of filters when the $N$ is large. Focusing on the $N<10$, the larger number of filters is, the higher the estimation accuracy. In contrast, when the $N$ is large, parts of the order are inversed. These results showed that $fils=3N$ was the best accuracy if enough memory can be allocated.

This result is not expected to occur with PE. Because PE trains each model independently, the performance of individual models is limited when the number of filters is few. On the other hand, EE trains each ensemble model with a single loss function, allowing each ensemble to acquire a different role.

\section{Conclusion}
In this study, we proposed EE as a method that is easy to implement and can be expected to improve accuracy by ensemble learning with a single model. In addition, as input variationer for EE, we proposed repeating input, using DAs, using modality, and masking. After proving that EE is equivalent to ME, which combines the inputs and outputs of multiple models, we experimentally demonstrated its equivalence and effectiveness using a benchmark dataset for sensor-based HAR. As a result, using a VGG architecture for the HASC dataset, whereas a single BL achieved 83.4\% mean accuracy, a simple EE ($fils=16$ and $N=4$) and tuned EE ($fils=3$ and $N=1024$) achieved 84.1\% and 85.3\% mean accuracies, respectively. Further, we experimentally analyzed the properties of EE and obtained the following.

\begin{itemize}
    \item When scaled up in terms of the number of filters, the accuracy curves of BL and EE showed roughly the same trend; however, when the number of ensembles was large, EE with a few number of filters outperformed EE with a large numbers of filters even if the total number of parameters is the same. Whereas the EE ($fils=16$ and $N=32$) achieved 84.6\% mean accuracy, the EE ($fils=3$ and $N=1024$) achieved 85.3\% mean accuracy for the HASC dataset.
    \item When scaled up in terms of the number of ensembles, the improvement in accuracy tended to continue without an upper limit in the range observed, and EE tended to outperform PE when the number of ensembles was larger than a certain value (about 0.2\% improvement for the HASC dataset.)
    \item Input variationer tended to improve the estimation accuracy when using data augmentation or modality ensemble.
\end{itemize}

    In this study, we clarified the effectiveness and characteristics of EE through the experiments in HAR field; however, EE can be applied to other fields (e.g., image recognition and signal processing.) One of the input variationer: modality ensemble is, especially for the HAR field when using multi-sensors. We will investigate the application of EE and other styles of input variationer in other filed as future work.

\appendix
\subsection{Acronyms definition}
We define all acronyms in this paper in Table \ref{table:acronyms}.
\begin{table}[t]
\caption{Acronyms definition in this paper}
\label{table:acronyms}
\hbox to\hsize{\hfil
\begin{tabular}{cc} \hline \hline
    acronyms &	definition \\ \hline
    HAR &	Human activity recognition \\ 
    IMU &	Intertial measurement unit \\ 
    CNN &	Convolutional neural network \\ 
    LSTM &	Long short-term memory \\ 
    GE &	Group ensemble \\ 
    BL &	Baseline \\ 
    PE &	Pure ensemble \\ 
    ME &	Merge ensemble \\ 
    EE &	Easy ensemble \\ 
    GAP &	Global average pooling \\ 
    DA &	Data augmentation \\ \hline        
\end{tabular}\hfil}
\end{table}

\subsection{Model details}
Table \ref{table:params} shows the hyperparameters, the number of trainable parameters, and floating-point operations (FLOPs) for each model in Section IV-B. Conv3 denotes the convolutional layer with the kernel size of 3, whose parameters are ``in channels,'' ``out channels,'' and ``groups,'' respectively. Norm denotes the normalization layer. MaxPool denotes the max pooling layer whose parameters are ``kernel size,'' ``stride,'' ``padding,'' and ``dilation,'' respectively. Output denotes the output layer consisting of the fully-connected (FC) layer whose parameters are ``in features'' and ``out features,'' respectively. ME is the same as four BLs, and their outputs are added, finally. The number of parameters and FLOPs are the same between ME and EE because they are entirely the same. 
\begin{table}[t]
    \newcolumntype{A}{>{\centering\arraybackslash}p{1em}}
    \newcolumntype{B}{>{\centering\arraybackslash}p{5.5em}}
    \newcolumntype{C}{>{\centering\arraybackslash}p{5em}}
\caption{Model hyperparameters which we used in experiments.}
\label{table:params}
\hbox to\hsize{\hfil
\begin{tabular}{Ac|B|Cc|B} \hline \hline
    stage &	layer &	BL &	\multicolumn{2}{c|}{ME} &	EE \\  \hline
    1 &	Conv3 &	3, 16, 1 &	3, 16, 1 &	\multirow{23}{*}{$\left. \begin{tabular}{@{}l@{}}
        \\ \\ \\ \\ \\ \\ \\ \\ \\ \\ \\ \\ \\ \\ \\ \\ \\ \\ \\ \\ \\ \\
        \end{tabular} \right\}$ $\times$ 4} &	12, 64, 4 \\ 
    1 &	Norm &	LayerNorm() &	LayerNorm() &	 &	GroupNorm(4) \\ 
    1 &	MaxPool &	2, 2, 0, 1 &	2, 2, 0, 1 &	 &	2, 2, 0, 1 \\  \cline{0-2} \cline{6-6}
    2 &	Conv3 &	16, 32, 1 &	16, 32, 1 &	 &	64, 128, 4 \\ 
    2 &	Norm &	LayerNorm() &	LayerNorm() &	 &	GroupNorm(4) \\ 
    2 &	MaxPool &	2, 2, 0, 1 &	2, 2, 0, 1 &	 &	2, 2, 0, 1 \\  \cline{0-2} \cline{6-6}
    3 &	Conv3 &	32, 64, 1 &	32, 64, 1 &	 &	128, 256, 4 \\ 
    3 &	Norm &	LayerNorm() &	LayerNorm() &	 &	GroupNorm(4) \\ 
    3 &	Conv3 &	64, 64, 1 &	64, 64, 1 &	 &	256, 256, 4 \\ 
    3 &	Norm &	LayerNorm() &	LayerNorm() &	 &	GroupNorm(4) \\ 
    3 &	MaxPool &	2, 2, 0, 1 &	2, 2, 0, 1 &	 &	2, 2, 0, 1 \\  \cline{0-2} \cline{6-6}
    4 &	Conv3 &	64, 128, 1 &	64, 128, 1 &	 &	256, 512, 4 \\ 
    4 &	Norm &	LayerNorm() &	LayerNorm() &	 &	GroupNorm(4) \\ 
    4 &	Conv3 &	128, 128, 1 &	128, 128, 1 &	 &	512, 512, 4 \\ 
    4 &	Norm &	LayerNorm() &	LayerNorm() &	 &	GroupNorm(4) \\ 
    4 &	MaxPool &	2, 2, 0, 1 &	2, 2, 0, 1 &	 &	2, 2, 0, 1 \\  \cline{0-2} \cline{6-6}
    5 &	Conv3 &	128, 128, 1 &	128, 128, 1 &	 &	512, 512, 4 \\ 
    5 &	Norm &	LayerNorm() &	LayerNorm() &	 &	GroupNorm(4) \\ 
    5 &	Conv3 &	128, 128, 1 &	128, 128, 1 &	 &	512, 512, 4 \\ 
    5 &	Norm &	LayerNorm() &	LayerNorm() &	 &	GroupNorm(4) \\ 
    5 &	MaxPool &	2, 2, 0, 1 &	2, 2, 0, 1 &	 &	2, 2, 0, 1 \\  \cline{0-2} \cline{6-6}
    6 &	Flatten &	GAP() &	GAP() &	 &	GAP() \\ 
    \multirow{2}{*}{6} &    \multirow{2}{*}{Output} &	\multirow{2}{*}{FC(128, 6)} &	FC(128, 6) &	 &	\multirow{2}{*}{FC(512, 6)} \\ 
     &	 &	 &	add() &	 &	 \\ \hline
    \multicolumn{2}{c|}{\# params.} &	$0.193 \times 10^6$ &	\multicolumn{2}{c|}{$0.772 \times 10^6$} &	$0.772 \times 10^6$ \\ \hline
    \multicolumn{2}{c|}{FLOPs} &	$5.42 \times 10^6$ &	\multicolumn{2}{c|}{$21.69 \times 10^6$} &	$21.69 \times 10^6$ \\ \hline 
\end{tabular}\hfil}
\end{table}

\bibliographystyle{IEEEtran}
\bibliography{mybib}

\begin{IEEEbiography}[{\includegraphics[width=1in,height=1.25in,clip,keepaspectratio]{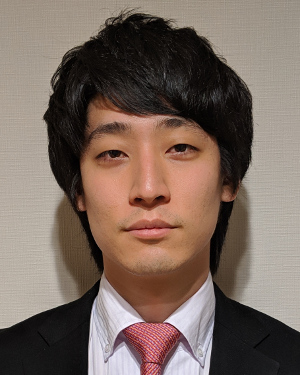}}]{Tatsuhito Hasegawa}
  (Member, IEEE) received the Ph.D. degree in engineering from Kanazawa University, Kanazawa, in 2015. From 2011 to 2013, he was a System Engineer with Fujitsu Hokuriku Systems Ltd. From 2014 to 2017, he was an Assistant with Tokyo Healthcare University. From 2017 to 2020, he was a Senior Lecturer with the Graduate School of Engineering, University of Fukui. He is currently an Associate Professor. His study interests include HAR, applying deep learning, and intelligent learning support system. He is also a member of IPSJ. 
\end{IEEEbiography}

\begin{IEEEbiography}[{\includegraphics[width=1in,height=1.25in,clip,keepaspectratio]{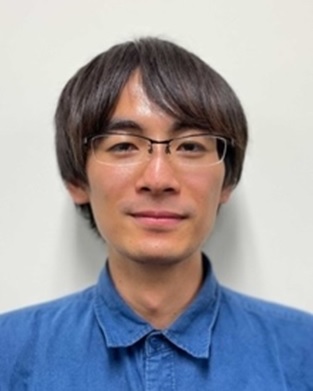}}]{Kazuma Kondo}
  received the M.S. degree in Engineering from the Graduate School of Engineering, University of Fukui, Japan, in 2022. He is currently working with NEC Solution Innovators, Ltd. His study interests include human activity recognition and deep learning application. 
\end{IEEEbiography}

\end{document}